\documentclass[sn-mathphys,Numbered]{sn-jnl}

\usepackage{graphicx}%
\usepackage{multirow}%
\usepackage{amsmath,amssymb,amsfonts}%
\usepackage{amsthm}%
\usepackage{mathrsfs}%
\usepackage[title]{appendix}%
\usepackage{xcolor}%
\usepackage{textcomp}%
\usepackage{manyfoot}%
\usepackage{booktabs}%
\usepackage{algorithm}%
\usepackage{algorithmicx}%
\usepackage{algpseudocode}%
\usepackage{listings}%
\usepackage{subfigure}
\usepackage{tablefootnote}

\theoremstyle{thmstyleone}%
\theoremstyle{thmstyletwo}%

\theoremstyle{thmstylethree}%

\begin{document}

\title[Article Title]{Human Action Recognition in Still Images Using ConViT}


\author[1]{\fnm{Seyed Rohollah} \sur{Hosseyni}}

\author*[1]{\fnm{Sanaz} \sur{Seyedin}}\email{sseyedin@aut.ac.ir}

\author[1]{\fnm{Hassan} \sur{Taheri}}


\affil[1]{\orgdiv{Department of Electrical Engineering}, \orgname{Amirkabir University of Technology (Tehran Polytechnic)}, \orgaddress{\street{350, Hafez Ave.}, \city{Tehran}, \country{Iran}}}


\abstract{Understanding the relationship between different parts of an image is crucial in a variety of applications, including object recognition, scene understanding, and image classification. Despite the fact that Convolutional Neural Networks (CNNs) have demonstrated impressive results in classifying and detecting objects, they lack the capability to extract the relationship between different parts of an image, which is a crucial factor in Human Action Recognition (HAR). To address this problem, this paper proposes a new module that functions like a convolutional layer that uses Vision Transformer (ViT). In the proposed model, the Vision Transformer can complement a convolutional neural network in a variety of tasks by helping it to effectively extract the relationship among various parts of an image. It is shown that the proposed model, compared to a simple CNN, can extract meaningful parts of an image and suppress the misleading parts.
The proposed model has been evaluated on the Stanford40 and PASCAL VOC 2012 action datasets and has achieved 95.5\% mean Average Precision (mAP) and 91.5\% mAP results, respectively, which are promising compared to other state-of-the-art methods.}




\maketitle

\section{Introduction}\label{sec1}
\begin{figure}[t]
	\centering     
	\subfigure[Blowing bubbles]{
		\label{spatial-introduction-b}
		\includegraphics[scale=0.3]{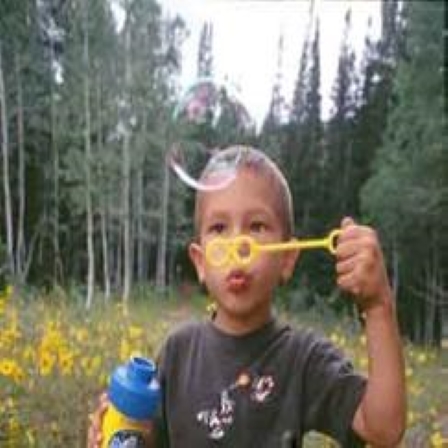}
	}
	\subfigure[Brushing teeth]{
		\label{spatial-introduction-c}
		\includegraphics[scale=0.3]{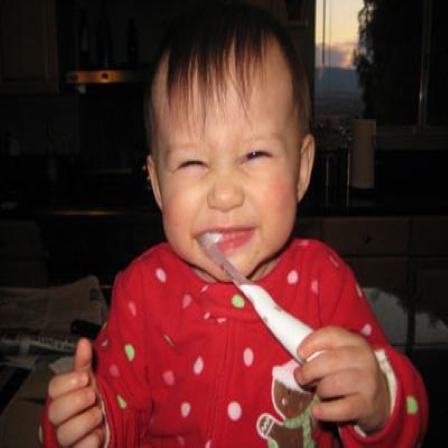}
	}
	\caption{  
		Inter-class similarity
	}
	\label{similar}
\end{figure}

\begin{figure}[t]
	\centering     
	\subfigure[Taking photos]{
		\label{spatial-introduction-b}
		\includegraphics[scale=0.3]{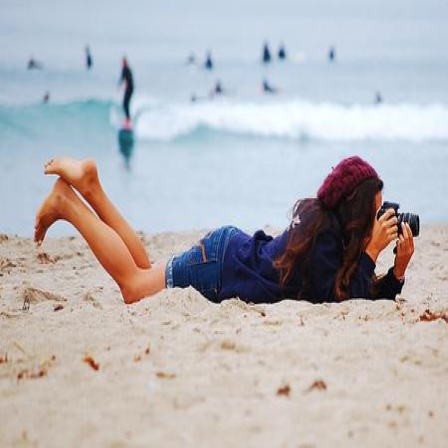}
	} 
	\subfigure[Taking photos]{
		\label{spatial-introduction-c}
		\includegraphics[scale=0.3]{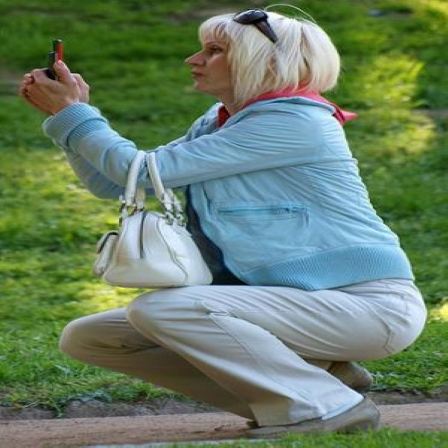}
	}
	\caption{  
		Intra-class difference
	}
	\label{diff}
\end{figure}

In recent years, human action recognition has become a topic that has attracted significant attention from researchers \cite{Guodongsurvey, 7916717, Vishwakarma}. One of the key factors that makes this topic significant is its relevance to real-life situations. There are many significant applications of human action recognition, such as intelligent surveillance \cite{6129539}, image annotations \cite{7780966}, healthcare \cite{ReviewonApplications}, image and video retrieval \cite{7916717}, human-object interaction \cite{799904}, and entertainment \cite{VALLIM20136258}. Human action recognition can be performed using either video-based or image-based techniques. Without temporal information, recognizing actions in images is harder due to the absence of motion information, which is only present in videos \cite{8965014}. Static information can provide insight into interactions between humans and objects, human pose, salient regions, objects, and background information \cite{AshrafiSeyed}. Recognizing actions in images poses several challenges such as dealing with unrelated objects and backgrounds, occluded backgrounds, intra-class variation, and inter-class similarities \cite{Chen2004}. As an illustration, Figure \ref{similar}  depicts two distinct categories and demonstrates comparable poses in both images. Figure \ref{diff} illustrates two equivalent classes that exhibit different poses and objects \cite{Chen2004}.

\begin{figure}[t]
	\centering     
	\subfigure[Fixing a bike]{
		\label{fdafdafa}
		\includegraphics[scale=0.3]{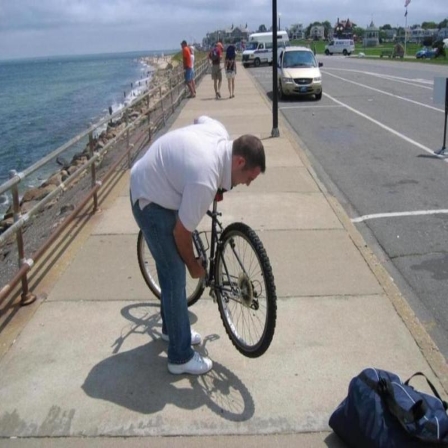}
	}
	\subfigure[Feeding a horse]{
		\label{fdafdafafdsfds}
		\includegraphics[scale=0.3]{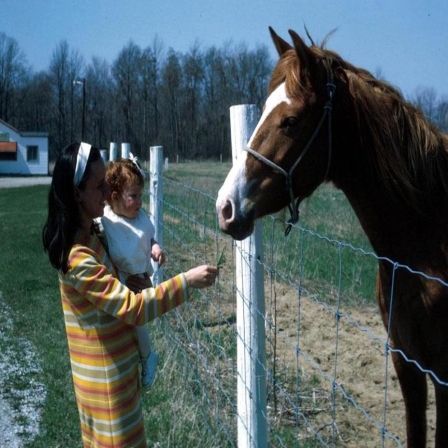}
	}
	\caption{  
		Importance of the relations between different areas of the image
	}
	\label{fdafdafafdsfdsfsdf}
\end{figure}

Over the past few years, deep convolutional neural networks (CNNs) \cite{7780459, 51456465465465} have demonstrated impressive results in various visual recognition tasks. These networks can extract features from images using convolution and downsampling operations and leverage their hierarchical structure to achieve high-level and discriminative representations \cite{6909618}. However, a limitation of these networks is their inability to extract the relationships between different regions of an image, which can hinder tasks like action recognition. In action recognition, it is crucial to extract the relationships between different regions of an image. For instance, as shown in Figure \ref{fdafdafa}, a failure to do so could lead to misclassification, such as mistaking a person repairing a bicycle for cycling. In Figure \ref{fdafdafafdsfds},  since the model has extracted horse and human information, if it  does not consider the relationship between regions in this image, it may incorrectly recognize it as a person riding a horse.

By utilizing self-attention mechanism, the Transformer model \cite{NIPS2017_3f5ee243} achieved impressive results in natural language processing, whereas the Vision Transformer \cite{vitjkljlfsd} applies the same technique to extract semantic relationships in images.

The ConViT model does not require additional information such as bounding boxes or keypoints and only relies on action labels. However, for complex datasets such as the PASCAL VOC 2012 Action dataset, which contains images with multiple individuals performing various actions, an object detection model is necessary, because assigning a single label to all individuals in an image is not practical

The ConViT model was evaluated on two publicly available datasets, the Stanford40 and PASCAL VOC 2012 Action datasets, achieving mAP (mean Average Precision) scores of 95.2\% and 87.9\%, respectively. On the Stanford40 dataset, the ConViT model outperformed other state-of-the-art methods. To further improve the model's performance on the PASCAL VOC 2012 Action dataset, a human classification branch can be added to leverage human spatial information. Integrating the ConViT model and the human classification branch improved the performance on both datasets, achieving mAP scores of 95.5\% and 91.5\%, respectively.

\section{Related Works}\label{sec2}
This section will be divided into two parts, the first of which will explore previous work on recognizing human actions in images, and the second of which will discuss attention-based methods.

\subsection{Action Recognition}
To recognize actions in images, researchers typically use pose and body part information, human-object interaction information, or a combination of both, in addition to scene information \cite{8214269, Chen2004}.

\textbf{Pose and body parts:} 
To use human body information such as pose and human body parts, additional information such as bounding box coordinates and keypoints are usually used\cite{8214269}. Thuran et al. \cite{Thurau2008PosePB} presented a way of recognizing human actions by using pose primitives. Bourdev et al. \cite{5459303} devised poselets, which are components that are closely grouped together in both configuration and appearance space. The deep version of poselets is employed by Gkioxari et al.\cite{7410641} for detecting and capturing parts of the human body under distinct poses, which are then fed into a CNN. Newell et al. \cite{Hourglass} developed a convolutional network architecture for the task of human pose estimation, which can be used for action recognition. The unified deep model proposed by Li et al. \cite{LI2020107341} explores the body structure information and combines multiple body structure cues to improve the robustness of action recognition in images.

\textbf{Human-Object interaction:} 
Yao et al. \cite{5540235} created a random field model to capture the mutual context of objects and human poses in activities involving human-object interaction. Desai et al. \cite{5543176} introduced an approach that models the contextual interaction between postured human bodies and surrounding objects. Gkioxari et al. \cite{Gkioxari2017DetectingAR} suggested a multi-branch model to detect human-object interaction by jointly detecting people and objects, which makes interaction inference efficient by merging the predictions. In another work, Gkioxari et al. \cite{7410486} employed an object detector called R-CNN to identify the location of objects, and then chose the most pertinent object for action recognition.

\textbf{Scene:} 
Delaitre et al. \cite{BMVC.24.97} examined the significance of the background scene context in action recognition and showed that it can enhance the performance of the task.  Zhang et al. \cite{7558119} employed an action mask to extract distinctive regions from a scene without the need for extra information like bounding boxes.

\subsection{Attention}
A crucial aspect of human perception is that we do not process an entire scene in its entirety all at once; rather, we selectively focus on parts of the visual scene to gather necessary information \cite{article524454}. The attention mechanism was first introduced for machine translation by Bahdanau et al. \cite{DzmitryBahdanau}. Different attention models have been employed in natural language processing \cite{9586824} and they have also been adapted to computer vision tasks.  Xu et al. \cite{Gkioxari2015ContextualLM} presented an attention-based model that can learn to automatically describe the content of images. To address fine-grained object classification, Zhao et al. \cite{7807286} proposed the diversified visual attention network (DVAN) that includes an attention mechanism. Transformer \cite{NIPS2017_3f5ee243} is a neural network that employs self-attention and has delivered remarkable outcomes on various NLP tasks such as machine translation, language modeling, and sentiment analysis. Vision Transformer (ViT) \cite{vitjkljlfsd} is the Transformer model designed for visual tasks.

\textbf{Attention in action recognition:}
A soft attention mechanism was added to the VGG16 model by Yan et al. \cite{8214269}, using two extra branches to capture global and local contextual information through scene-level and region-level attention.  SAAM-Nets, a multi-stage deep learning method proposed by Zheng et al. \cite{Chen2004}, employs spatial attention to extract action-specific visual semantic parts without requiring additional annotations.  Ashrafi et al. \cite{AshrafiSeyed} developed a model for recognizing actions that detects multiple salient regions using a multi-attention guided network in a weakly-supervised manner.  Ashrafi et al. \cite{ashrafi2023}  simultaneously employs body joint and object cues, integrating their features using a proposed attention module. This module, comprising dual self-attentions and a cross-attention, captures interactions among objects, joints, and their combinations. Bas et al.\cite{BAS2022116664} proposed a model that concurrently identifies action-related components within images, generates attention masks for both pixel and image-level action cues, and predicts image-level action labels without the need for explicit detectors.

 \section{Approach} \label{section-approach}
The proposed ConViT model for recognizing human action in still images is introduced in this section, along with a human classification branch that enhances the model's performance.
\subsection{Overview}
Spatial features are typically obtained through convolutional networks in action recognition systems. These features are then utilized to obtain other information that can enhance the model's performance. Figure \ref{model} illustrates the proposed model. The model comprises two parts. The first part is a fully convolutional network that uses ResNet50 to extract spatial features. The second part of the model includes two Vision Transformers. The purpose of using Vision Transformers in this research is to extract the relationship between different areas of an image, similar to the way transformers are employed in NLP studies. The feature map obtained from the convolutional neural network undergoes two Vision Transformers to generate a new feature map that encodes information about the relationships between the pixels. The final step involves applying the Global Average Pooling (GAP) operator to the new feature map, passing the output feature vector through a fully connected (fc) layer, and predicting class scores. The proposed model is called ConViT.

\subsection{CNN}
The initial component in the proposed model is a fully convolutional network, such as ResNet50, although other convolutional networks like VGG can also be utilized in this segment, as its aim is to extract spatial information from the image. Each pixel in the final feature map of the convolutional network represents a specific region of the image.

\begin{figure}[h]
	\centering
	\includegraphics[scale=0.24]{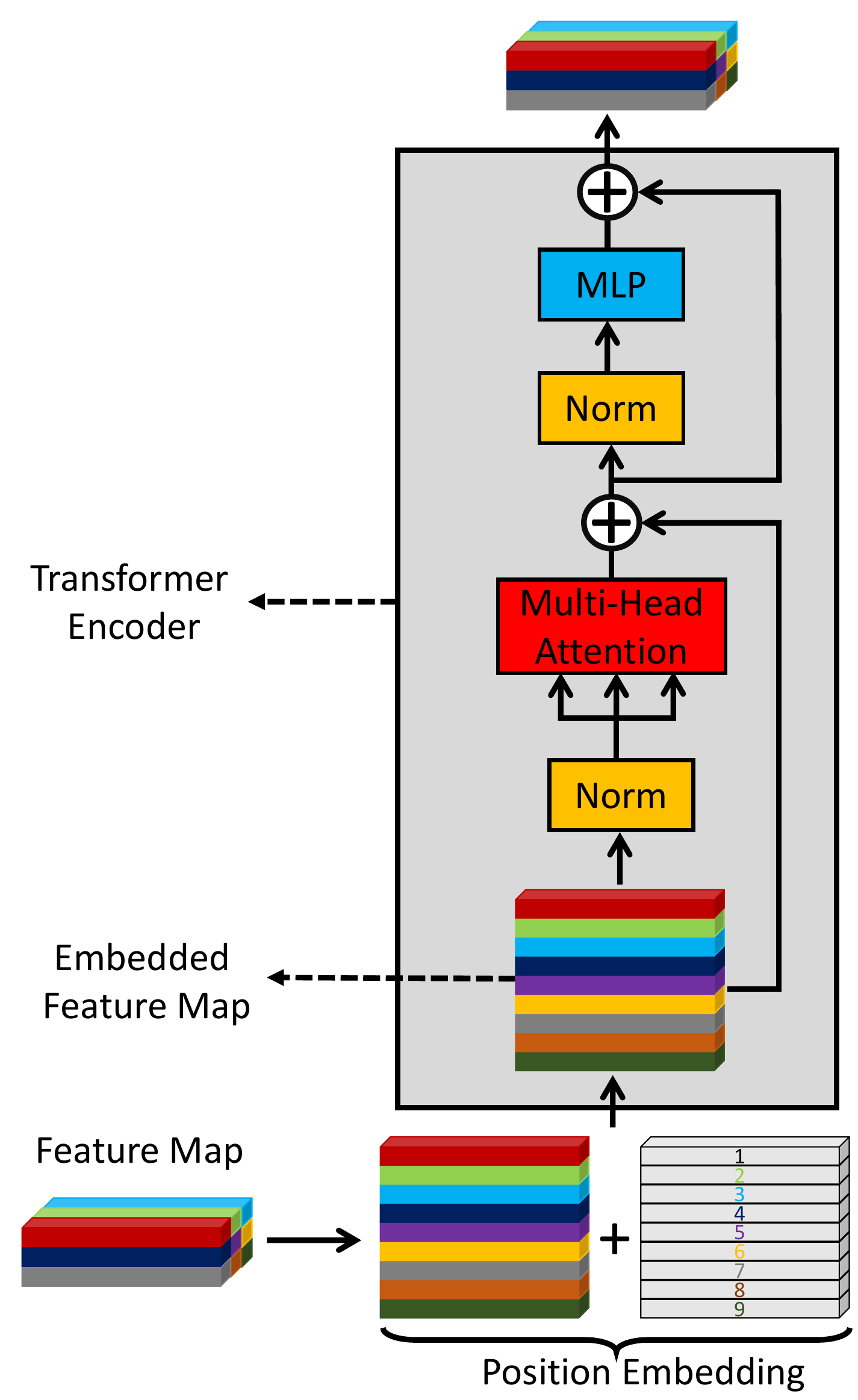}
	\caption{The Modified ViT takes a feature map as input and produces a new feature map as output, which contains information about the relationships between different regions of the image.}
	\label{my-encoder}
\end{figure}

\subsection{Modified ViT}
The second part of the proposed model involves using ViT to extract relationships between different regions of the image.

The feature map obtained from ResNet can be fed into ViT for further processing. To preserve the positional information of the pixels, position embedding is used initially. Unlike the original ViT model \cite{vitjkljlfsd}, where the pixels of each image region are converted to a vector and linearly mapped, the feature vectors for each pixel are directly fed into ViT without any modifications, as shown in Figure \ref{my-encoder}. The proposed ViT model does not use a class token next to different image regions as input, unlike the original ViT model \cite{vitjkljlfsd}. The proposed ViT model functions similarly to a convolutional layer, taking a feature map as input and generating a feature map as output, whereas the original ViT model takes raw images as input and produces class scores as output.

\begin{figure}[th]
	\centering
	\includegraphics[width=0.8\textwidth]{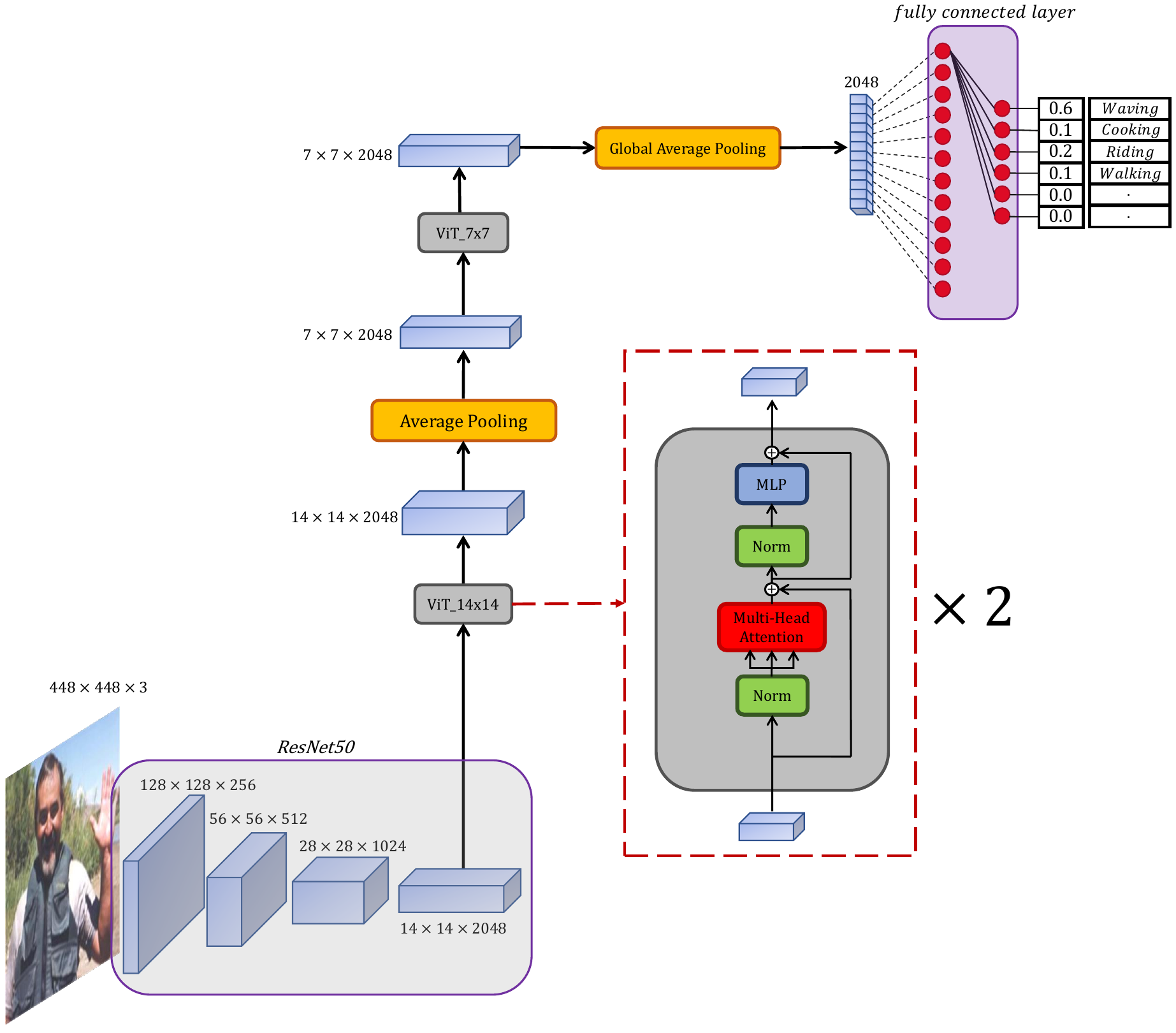}
	\caption{ConViT model architecture. Our proposed ConViT model consists of two parts. First, the input image is passed through a CNN, which generates a feature map capturing high-level spatial features. Second, this feature map is processed by two Vision Transformers (ViTs) that are capable of learning semantic relationships between different regions of the image.} 
	\label{model}
\end{figure}


The input to the first ViT in Figure \ref{model} is a feature map with dimensions of $14 \times 14 \times 2048$. Since this ViT receives 196 input regions, establishing relationships between them can be challenging due to the high number of areas. Therefore, the initial ViT model may have difficulty effectively extracting the relationships between different regions of the image, resulting in less attention being paid to discriminative regions and more to misleading ones. To address this issue, a second ViT is employed with a feature map of dimensions $7 \times 7 \times 2048$ as input. Since the number of areas is reduced to 49, the second ViT can more easily learn the relationships between these areas. Section \ref{ablation-section} will demonstrate how the second ViT can selectively attend to informative regions that the first ViT missed while disregarding misleading regions.

\subsection{Human Classification Branch}
\begin{figure}[h]
	\centering
	\includegraphics[width=0.55\textwidth]{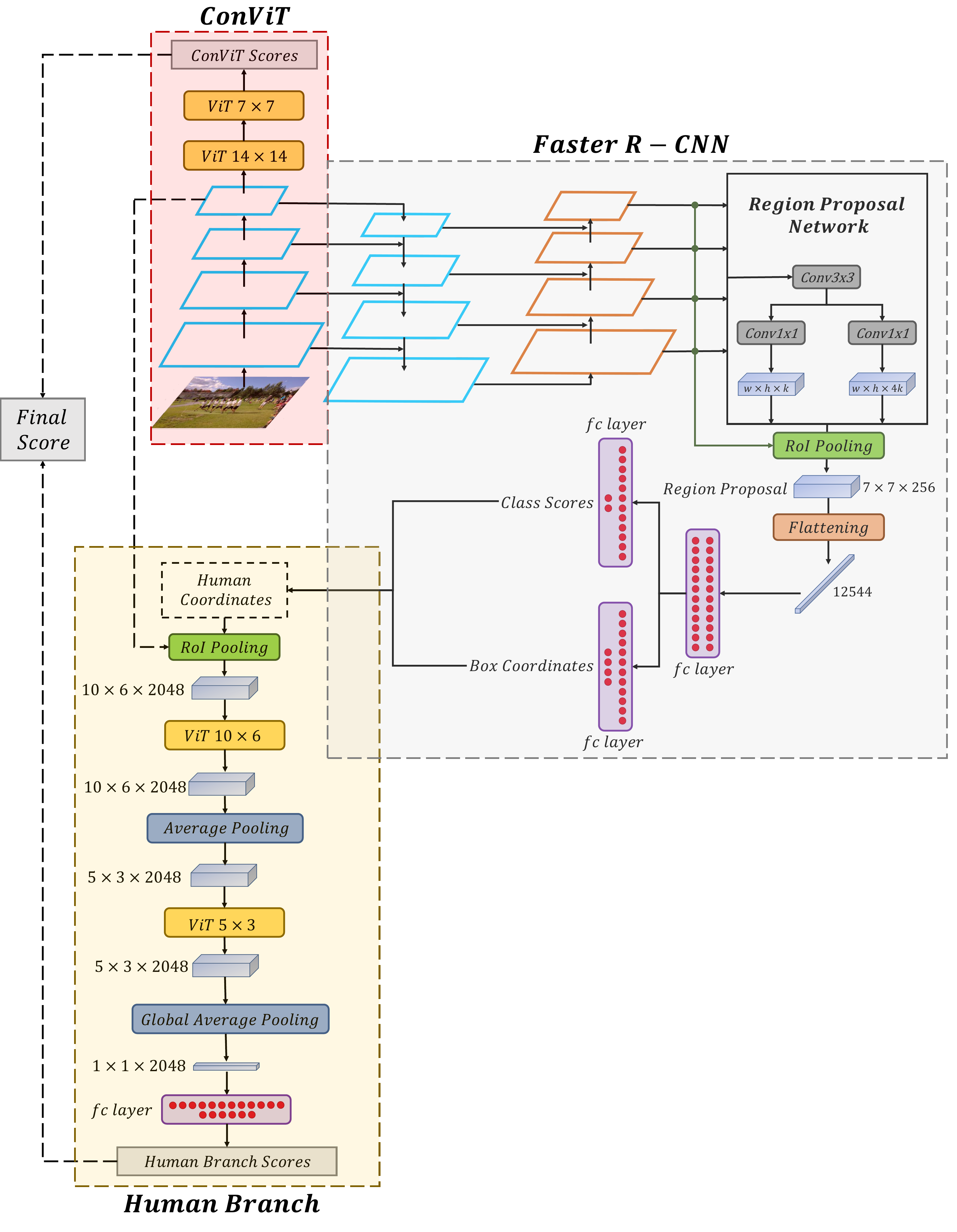}
	\caption{Model architecture. It's shown that once Faster R-CNN provides human bounding box coordinates, these coordinates are used to extract the feature map of the corresponding person from the final ResNet50 feature map. Subsequently, this cropped feature map goes through two ViTs. Finally, the predicted scores from this branch are combined with ConViT's predicted scores, each being appropriately weighted.} 
	\label{human-branch-details}
\end{figure}

ConViT incorporates all contextual details in the scene, which can cause the network to be misguided by the presence of irrelevant objects or background. Additionally, the ConViT model provides only one prediction for the entire image, whereas the input image can contain multiple people performing diverse actions. Hence, it would be advantageous to incorporate an additional branch that only considers the target person's area and predicts the action scores for each person. To extract the target person's area from the image, an object detection model is required. In this study, we employ the Faster R-CNN model \cite{NIPS2015_14bfa6bb} with PANet \cite{8579011} in its Neck as shown in Figure \ref{human-branch-details}.

Using RoI Pooling and the coordinates of the target person predicted by Faster R-CNN, the target person's area can be obtained from the latest feature map produced by ResNet50, which has dimensions of $10 \times 6 \times 2048$. Then, we pass this feature map to two dedicated ViT models, ViT 10x6 and ViT 5x3, which are meant for the classification of humans. Finally, a new feature map with dimensions of $5 \times 3 \times 2048$ is produced, and class scores are predicted.

\subsection{Final Prediction}
The ConViT and human branch have almost different designs and undergo distinct input training. Consequently, the impact of their outputs will vary depending on the actions\cite{Chen2004}. To be more specific, the ConViT model utilizes the complete image as an input to gather global information, while the human classification branch focuses on the target person's area to extract local information. The ConViT and human branch scores are combined to obtain the final prediction score as follows:
\begin{equation}
	P_{final} = W_{ConViT} \times P_{ConViT} + W_{human} \times P_{human}
\end{equation}
The weights $W_{ConViT}$ and $W_{human}$, which are responsible for the prediction scores of ConViT and the human branch, are selected in a way that minimizes the loss on their respective datasets.

 \section{Experiments}
Our proposed method is evaluated on two action recognition datasets in this section, and subsequently, the results of the experiments on these datasets are analyzed.

\subsection{Datasets}
\textbf{Stanford40 dataset:} \cite{VALLIM20136258fdsfdsfsdf} This dataset consists of 40 human actions that are performed in daily life, such as reading books, riding, and cleaning the floor. It contains 9,532 images, which are divided into two sets: a training set with 4,000 images and a test set with 5,352 images. There are 180 to 300 images per action class.

\textbf{PASCAL VOC 2012 Action dataset:} \cite{EveringhamGWWZ10} In this dataset, there are 10 different action categories, with the training set containing 2296 images and the validation set having 2292 images. The test set is comprised of 4569 images and is used to evaluate our proposed method, which is trained on all images from the training and validation sets.

\subsection{Training Details}
The paper employs the following data augmentation techniques: cropping, flipping the image horizontally, mixup\cite{mixup}, and RandAugment\cite{RandAugment}. Initially, the images are resized to $448 \times 448 \times 3$, and a random crop is applied to the input image while preserving 70\% of the human bounding box in the cropped image. The image is then subjected to horizontal flipping, mixup, and RandAugment techniques. The value of $\alpha$ for mixup data augmentation is set to 0.4 \cite{RandAugment}.

First, the ConViT network is trained, with ResNet50 serving as its convolutional part, pre-trained on the ImageNet dataset \cite{ImageNet}. Once ConViT is trained, its weights are frozen, and training of Faster R-CNN is initiated. Faster R-CNN employs ResNet50 from ConViT as its backbone, and bounding box information is utilized during its training. Finally, training of the human classification branch begins with all weights of ConViT and Faster R-CNN fixed.

The proposed method is implemented using the PyTorch framework\cite{NEURIPS2019_9015}. During training, the Stochastic Gradient Descent optimizer is applied utilizing a learning rate set at 0.001, a momentum value of 0.9, and a weight decay factor of 3e-5. At the beginning, the learning rate is 0.001, and it undergoes a 0.1-fold reduction after every 10-epoch interval. We use a mini-batch of 16 per iteration. Each ViT has a depth of 2 and 4 heads.

\subsection{Comparison with Existing Methods}
\textbf{Stanford40 dataset:}
ConViT's performance on the Stanford40 test set is shown in Table \ref{stanford-convit}, indicating that it outperforms all previous approaches. Moreover, ConViT, in contrast to the other models which used for comparison, doesn't need any supplementary data or auxiliary algorithms like human bounding boxes or R-CNN during evaluation. The combination of ConViT and human branch scores yields an increase in the mAP value by 0.3\% compared to ConViT alone. The final prediction score is calculated as follows in this configuration:
\begin{equation}
	P_{final} = 0.83 \times P_{ConViT} + 0.17 \times P_{human}
\end{equation}

The train loss is minimized to derive the weights associated with ConViT and the human branch. To achieve this, we combined the weights of these two components to equal 1. Through experimentation with different weight values, we found that the optimal configuration for the Stanford40 train dataset was to allocate 0.83 weight to ConViT and 0.17 weight to the human branch. These determined weights were then employed to evaluate the model's performance on the Stanford40 test dataset. Evidently, the significance of ConViT surpasses that of the human branch, underscoring ConViT's importance in the Stanford40 dataset. The dataset comprises single-labeled images, with relatively few distinct individuals per image. Consequently, the utility of the human branch in this context is minimal, as reflected by its relatively diminutive weight.

\begin{table}[h]
	\centering
	\begin{tabular}{lc}
		\hline
		\multicolumn{1}{l}{Method}                               &  {mAP (\%)}  \\ \hline
		{Body Structure Cues}     \cite{LI2020107341}        & $93.8$          \\ 
		{R*CNN}                      \cite{7410486}          & $90.9$          \\ 
		{Multi-Attention Guided Network}  \cite{AshrafiSeyed}& $94.2$          \\ 
		{ResNet-50}         \cite{7780459}                & $87.2$          \\ 
		{SAAM-Nets}               \cite{Chen2004}            & $93.0$          \\ 
		{Multi-Branch Attention}         \cite{8214269}      & $90.7$         \\ 
		{MASPP}         \cite{ashrafi2023}      & $ 94.8$         \\ 
		{Top-down + Bottom-up Attention}         \cite{BAS2022116664}      & $ 91.0$         \\ 
		{Human-Object Relation Network}    \cite{9102933}     & $94.6$          \\ \hline
		{ConViT}     (Ours)   & $95.2$          \\ 
		{Human Branch}     (Ours)    & $93.1$          \\ 
		{ConViT + Human Branch}         (Ours)                         & $95.5$          \\ \hline
	\end{tabular}
	\caption{Evaluating the suggested approach in relation to previous studies on the Stanford40 test dataset.}
	\label{stanford-convit}
\end{table}

\begin{figure*}[th]
	\centering
	\includegraphics[width=0.95\textwidth]{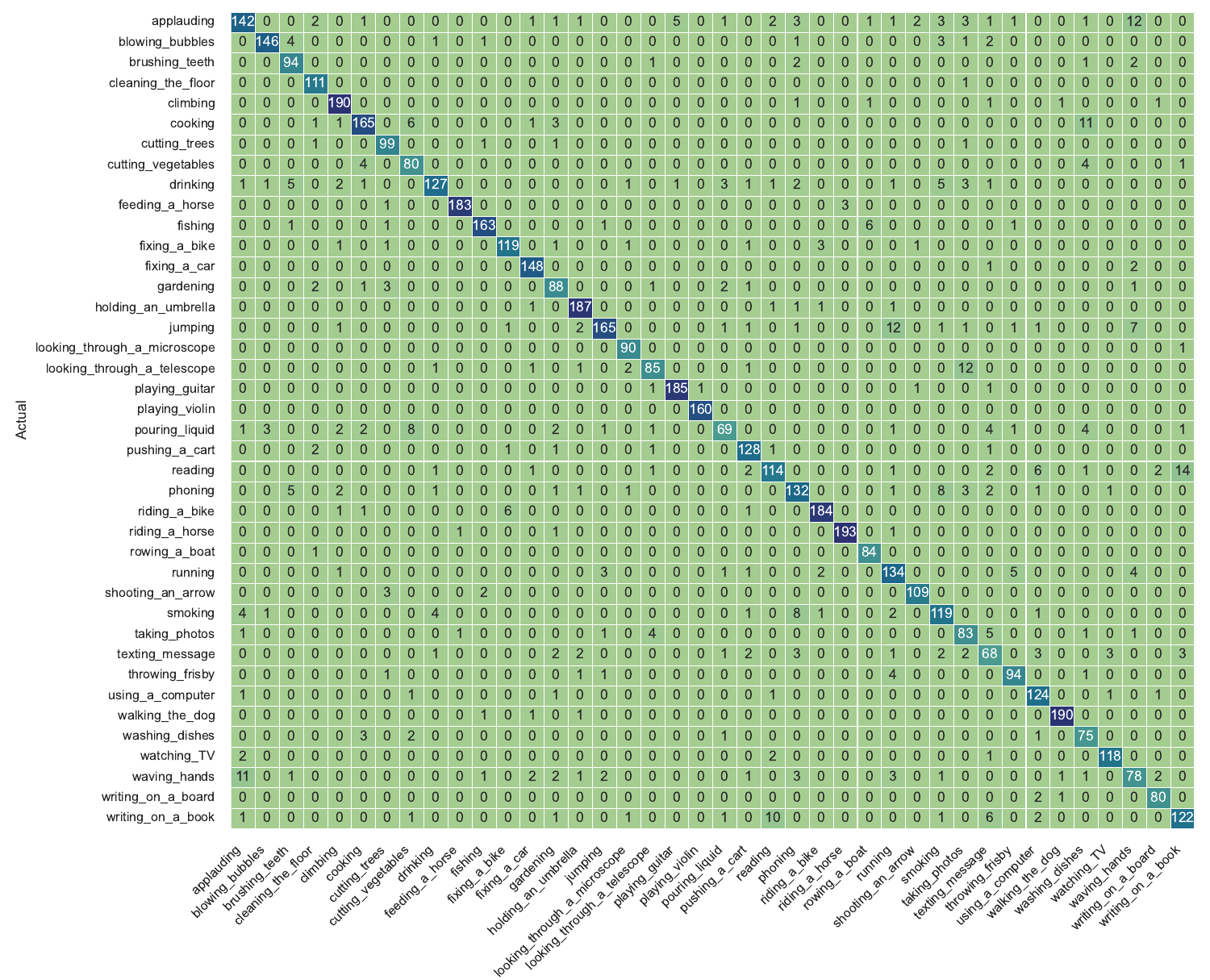}
	\caption{The confusion matrix for the proposed model on the Stanford-40 dataset.} 
	\label{stanford_confusion_matrix}
\end{figure*}

\textbf{PASCAL VOC 2012 Action dataset:}
The results on the PASCAL VOC 2012 action test set are presented in Table \ref{pascal-results}, indicating that ConViT did not perform well on this dataset due to its complexity. To address this issue, a human branch was added, which achieved an mAP value of 2.7\% higher than ConViT. When ConViT and human branch scores are fused, the final model performs 0.9\% better than the human branch alone, and the final scores are obtained using the following equation:
\begin{equation}
	P_{final} = 0.3 \times P_{ConViT} + 0.7 \times P_{human}
\end{equation}\label{eq-pascal}

The optimal weight values for ConViT and the human branch concerning the PASCAL VOC dataset were derived through minimizing its train loss. These weights were assigned while ensuring their sum equates to 1. Following trials with different weight combinations, it was established that ConViT's weight should be 0.3, while the human branch should be allocated a weight of 0.7. Ultimately, these derived weights were utilized to evaluate  the model's performance on the PASCAL VOC test dataset. It is apparent that in the context of the PASCAL VOC dataset, the weight assigned to the human branch exceeds that of ConViT, underscoring the pivotal role played by this branch within this dataset. This prominence stems from the dataset's characteristic inclusion of multi-labeled images, wherein each image can feature multiple individuals bearing distinct labels. Consequently, the inclusion of a human branch in such scenarios is notably impactful and contributes significantly to enhancing model performance, as evidenced by the remarkable improvements highlighted in Table \ref{pascal-results}.

\begin{table}[h]
	\centering
	\begin{tabular}{lc}
		\hline
		Method          &  {mAP (\%)} \\ \hline
		{Body Structure Cues}   \cite{LI2020107341}   & $93.5$        \\
		{R*CNN}     \cite{7410486}   & $90.2$        \\
		{ResNet-50}  \cite{7780459}     & $83.8$        \\
		{SAAM-Nets}   \cite{Chen2004}    & $84.8$        \\
		{Multi-Branch Attention}  (with bounding box)  \cite{8214269}  & $90.2$        \\
		{Multi-Branch Attention} (without bonding box) \cite{8214269}  & $84.5$        \\
		{MASPP}         \cite{ashrafi2023}      & $ 93.2$         \\ 
		{Top-down + Bottom-up Attention}         \cite{BAS2022116664}      & \hspace{0.1cm} $ 95.0$  \tablefootnote{This mAP is based on the PASCAL VOC validation set.}       \\ 
		{Human-Object Relation Network}     \cite{9102933}    & $92.8$        \\ \hline
		{ConViT}   (Ours)   & $87.9$        \\
		{Human Branch}     (Ours)    & $90.6$        \\
		{ConViT + Human Branch}        (Ours)                                      & 91.5       \\ \hline
	\end{tabular}
	\caption{ Evaluating the suggested approach in relation to previous studies on the PASCAL VOC 2012 action test set.}
	\label{pascal-results}
\end{table}

\subsection{Ablation Study and Visualization} \label{ablation-section}
In order to show the efficiency of each element in our approach, we employ ResNet50 as the backbone network in various configurations.  The results are shown in Table \ref{ablation}.

\begin{table}[h]
	\centering
	\begin{tabular}{lll}
		\hline
		Model & Pascal VOC  (mAP\%) &  Stanford40   (mAP\%) \\\hline
		Resnet50              & 86.2 &  93.0 \\ 
		ConViT                & 87.6 &  95.2 \\
		Human Branch          & 89.0 &   93.1  \\
		ConViT + Human Branch & 90.0  &  95.5  \\ \hline
	\end{tabular}
	\caption{Ablation study on PASCAL VOC 2012 action validation set and Stanford40 test set.}
	\label{ablation}
\end{table}

According to Table \ref{ablation}, model performance can be enhanced by the addition of two ViTs to ResNet50, thereby creating the ConViT model. ConViT employs ViTs to identify discriminative regions and suppress misleading regions. 

\begin{figure*}[th]
	\centering
	\includegraphics[width=0.95\textwidth]{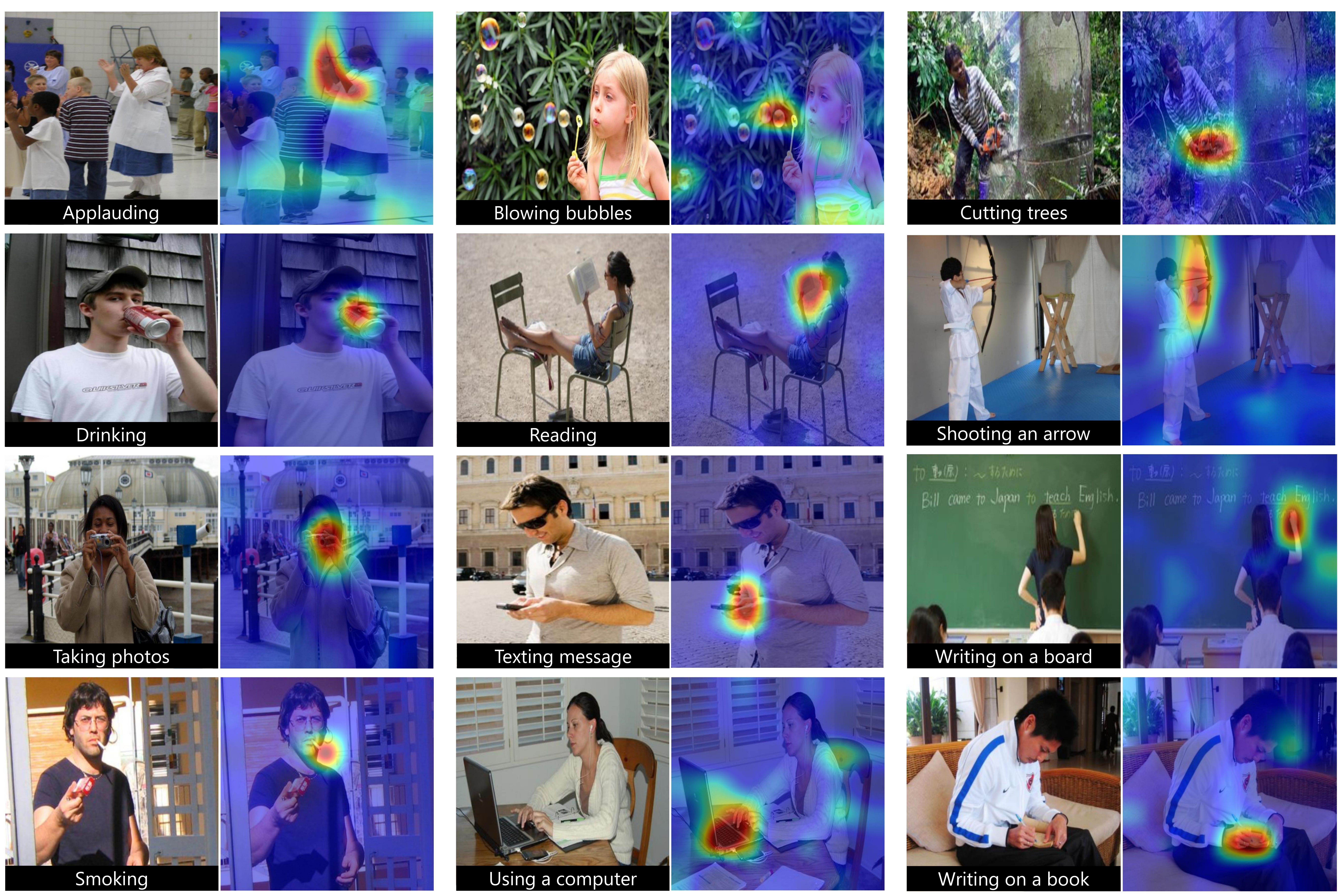}
	\caption{Grad-Cam visualization for some samples from Stanford40 dataset.}
	\label{Grad-Cam}
\end{figure*}

Figure \ref{Grad-Cam} presents a visual representation of Grad-Cam \cite{8237336} associated with the feature map produced by ViT 7x7, which is the final feature map of ConViT. Grad-Cam can be used to analyze the final feature map of convolutional models in terms of gradient information. According to figure \ref{Grad-Cam},  the ConViT model successfully detects discriminative regions corresponding to each class while ignoring misleading areas and backgrounds that do not contribute to the accurate classification.

\begin{figure*}[th]
	\centering
	\includegraphics[width=0.95\textwidth]{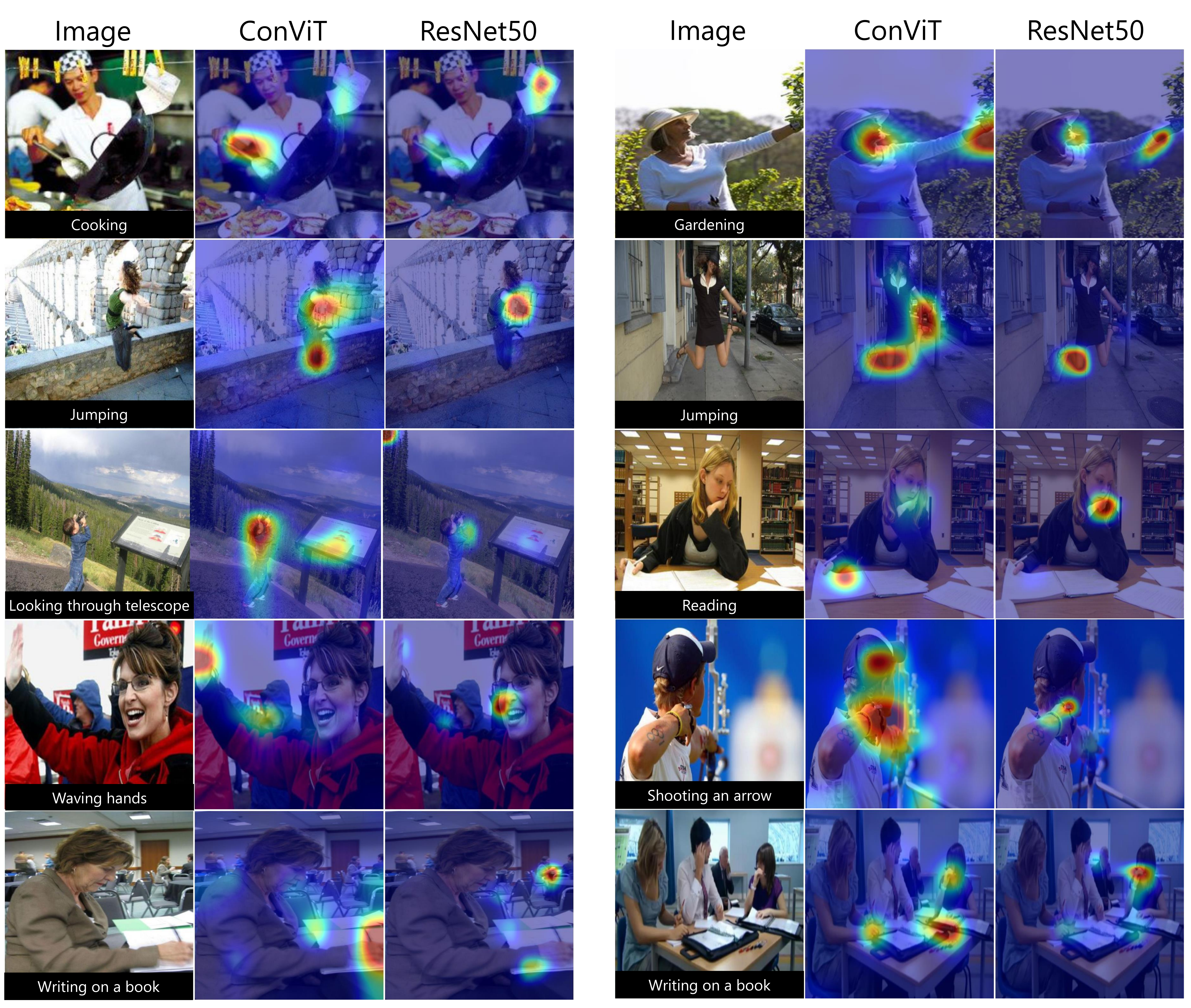}
	\caption{Comparing the final feature maps of ConViT and ResNet50 for samples on which ConViT made correct predictions while ResNet50 predicted incorrectly.} 
	\label{vit_vs_res}
\end{figure*}

The heatmaps corresponding to the final feature maps of both the ConViT and ResNet50 models are depicted in the figure \ref{vit_vs_res}. The ConViT and original ResNet50\cite{7780459} models were trained independently on the Stanford40 dataset. Subsequently, the final feature map generated by the ConViT model, through ViT 7x7, was utilized to compute the ConViT heatmap. Similarly, the final feature map of the original ResNet50 model was employed to compute the ResNet50 heatmap.  In this context, a heatmap denotes the outcome of averaging across the depth of feature map channels. This process results in an image size of 14x14 for ResNet50 and 7x7 for ConViT, each with 1 channel, which are then upscaled to match the size of the original image.

Figure \ref{vit_vs_res} shows that when ResNet50 couldn't find discriminative  regions  in certain input images, ConViT did an excellent job at finding those regions. For instance, in images labeled as "waving hands," "writing on book," and "Reading," ResNet50 exhibited incorrect recognition of vital areas, whereas ConViT adeptly pinpointed these areas and made accurate class predictions.

\section{Conclusion}
The ConViT model is proposed in this study for recognizing human actions in still images, consisting of two parts: CNN and ViT. CNN extracts high-level image features, while the proposed ViT learns the relationship between different areas of the image, which can work as a convolutional layer and be integrated with any CNN model. ConViT outperforms the current state-of-the-art models in Stanford40 dataset. For more challenging datasets, like PASCAL VOC 2012 action dataset, the model's performance is enhanced by embedding a human classification branch. The effectiveness of the human branch is directly linked to the performance of the object detector model in detecting human actions in complex datasets like PASCAL VOC 2012. \cite{7780966}

\bibliography{sn-bibliography}


\begin{thebibliography}{45}
\ifx \bisbn   \undefined \def \bisbn  #1{ISBN #1}\fi
\ifx \binits  \undefined \def \binits#1{#1}\fi
\ifx \bauthor  \undefined \def \bauthor#1{#1}\fi
\ifx \batitle  \undefined \def \batitle#1{#1}\fi
\ifx \bjtitle  \undefined \def \bjtitle#1{#1}\fi
\ifx \bvolume  \undefined \def \bvolume#1{\textbf{#1}}\fi
\ifx \byear  \undefined \def \byear#1{#1}\fi
\ifx \bissue  \undefined \def \bissue#1{#1}\fi
\ifx \bfpage  \undefined \def \bfpage#1{#1}\fi
\ifx \blpage  \undefined \def \blpage #1{#1}\fi
\ifx \burl  \undefined \def \burl#1{\textsf{#1}}\fi
\ifx \doiurl  \undefined \def \doiurl#1{\url{https://doi.org/#1}}\fi
\ifx \betal  \undefined \def \betal{\textit{et al.}}\fi
\ifx \binstitute  \undefined \def \binstitute#1{#1}\fi
\ifx \binstitutionaled  \undefined \def \binstitutionaled#1{#1}\fi
\ifx \bctitle  \undefined \def \bctitle#1{#1}\fi
\ifx \beditor  \undefined \def \beditor#1{#1}\fi
\ifx \bpublisher  \undefined \def \bpublisher#1{#1}\fi
\ifx \bbtitle  \undefined \def \bbtitle#1{#1}\fi
\ifx \bedition  \undefined \def \bedition#1{#1}\fi
\ifx \bseriesno  \undefined \def \bseriesno#1{#1}\fi
\ifx \blocation  \undefined \def \blocation#1{#1}\fi
\ifx \bsertitle  \undefined \def \bsertitle#1{#1}\fi
\ifx \bsnm \undefined \def \bsnm#1{#1}\fi
\ifx \bsuffix \undefined \def \bsuffix#1{#1}\fi
\ifx \bparticle \undefined \def \bparticle#1{#1}\fi
\ifx \barticle \undefined \def \barticle#1{#1}\fi
\bibcommenthead
\ifx \bconfdate \undefined \def \bconfdate #1{#1}\fi
\ifx \botherref \undefined \def \botherref #1{#1}\fi
\ifx \url \undefined \def \url#1{\textsf{#1}}\fi
\ifx \bchapter \undefined \def \bchapter#1{#1}\fi
\ifx \bbook \undefined \def \bbook#1{#1}\fi
\ifx \bcomment \undefined \def \bcomment#1{#1}\fi
\ifx \oauthor \undefined \def \oauthor#1{#1}\fi
\ifx \citeauthoryear \undefined \def \citeauthoryear#1{#1}\fi
\ifx \endbibitem  \undefined \def \endbibitem {}\fi
\ifx \bconflocation  \undefined \def \bconflocation#1{#1}\fi
\ifx \arxivurl  \undefined \def \arxivurl#1{\textsf{#1}}\fi
\csname PreBibitemsHook\endcsname

\bibitem[\protect\citeauthoryear{Guo and Lai}{2014}]{Guodongsurvey}
\begin{barticle}
\bauthor{\bsnm{Guo}, \binits{G.}},
\bauthor{\bsnm{Lai}, \binits{A.}}:
\batitle{A survey on still image based human action recognition}.
\bjtitle{Pattern Recognition}
\bvolume{47}(\bissue{10}),
\bfpage{3343}--\blpage{3361}
(\byear{2014})
\doiurl{10.1016/j.patcog.2014.04.018}
\end{barticle}
\endbibitem

\bibitem[\protect\citeauthoryear{Dhamsania and Ratanpara}{2016}]{7916717}
\begin{bchapter}
\bauthor{\bsnm{Dhamsania}, \binits{C.J.}},
\bauthor{\bsnm{Ratanpara}, \binits{T.V.}}:
\bctitle{A survey on human action recognition from videos}.
In: \bbtitle{2016 Online International Conference on Green Engineering and
  Technologies (IC-GET)},
pp. \bfpage{1}--\blpage{5}
(\byear{2016}).
\doiurl{10.1109/GET.2016.7916717}
\end{bchapter}
\endbibitem

\bibitem[\protect\citeauthoryear{Vishwakarma and Agrawal}{2012}]{Vishwakarma}
\begin{botherref}
\oauthor{\bsnm{Vishwakarma}, \binits{S.}},
\oauthor{\bsnm{Agrawal}, \binits{A.}}:
A survey on activity recognition and behavior understanding in video
  surveillance.
The Visual Computer
\textbf{29}
(2012)
\doiurl{10.1007/s00371-012-0752-6}
\end{botherref}
\endbibitem

\bibitem[\protect\citeauthoryear{Popoola and Wang}{2012}]{6129539}
\begin{barticle}
\bauthor{\bsnm{Popoola}, \binits{O.P.}},
\bauthor{\bsnm{Wang}, \binits{K.}}:
\batitle{Video-based abnormal human behavior recognition—a review}.
\bjtitle{IEEE Transactions on Systems, Man, and Cybernetics, Part C
  (Applications and Reviews)}
\bvolume{42}(\bissue{6}),
\bfpage{865}--\blpage{878}
(\byear{2012})
\doiurl{10.1109/TSMCC.2011.2178594}
\end{barticle}
\endbibitem

\bibitem[\protect\citeauthoryear{Yatskar et~al.}{2016}]{7780966}
\begin{bchapter}
\bauthor{\bsnm{Yatskar}, \binits{M.}},
\bauthor{\bsnm{Zettlemoyer}, \binits{L.}},
\bauthor{\bsnm{Farhadi}, \binits{A.}}:
\bctitle{Situation recognition: Visual semantic role labeling for image
  understanding}.
In: \bbtitle{2016 IEEE Conference on Computer Vision and Pattern Recognition
  (CVPR)},
pp. \bfpage{5534}--\blpage{5542}
(\byear{2016}).
\doiurl{10.1109/CVPR.2016.597}
\end{bchapter}
\endbibitem

\bibitem[\protect\citeauthoryear{Ranasinghe
  et~al.}{2016}]{ReviewonApplications}
\begin{botherref}
\oauthor{\bsnm{Ranasinghe}, \binits{S.}},
\oauthor{\bsnm{Al~Machot}, \binits{F.}},
\oauthor{\bsnm{Mayr}, \binits{H.}}:
A review on applications of activity recognition systems with regard to
  performance and evaluation.
International Journal of Distributed Sensor Networks
\textbf{12}
(2016)
\doiurl{10.1177/1550147716665520}
\end{botherref}
\endbibitem

\bibitem[\protect\citeauthoryear{Lee and Kim}{1999}]{799904}
\begin{barticle}
\bauthor{\bsnm{Lee}, \binits{H.-K.}},
\bauthor{\bsnm{Kim}, \binits{J.H.}}:
\batitle{An hmm-based threshold model approach for gesture recognition}.
\bjtitle{IEEE Transactions on Pattern Analysis and Machine Intelligence}
\bvolume{21}(\bissue{10}),
\bfpage{961}--\blpage{973}
(\byear{1999})
\doiurl{10.1109/34.799904}
\end{barticle}
\endbibitem

\bibitem[\protect\citeauthoryear{Vallim et~al.}{2013}]{VALLIM20136258}
\begin{barticle}
\bauthor{\bsnm{Vallim}, \binits{R.M.M.}},
\bauthor{\bsnm{{Andrade Filho}}, \binits{J.A.}},
\bauthor{\bsnm{{de Mello}}, \binits{R.F.}},
\bauthor{\bsnm{{de Carvalho}}, \binits{A.C.P.L.F.}}:
\batitle{Online behavior change detection in computer games}.
\bjtitle{Expert Systems with Applications}
\bvolume{40}(\bissue{16}),
\bfpage{6258}--\blpage{6265}
(\byear{2013})
\doiurl{10.1016/j.eswa.2013.05.059}
\end{barticle}
\endbibitem

\bibitem[\protect\citeauthoryear{Mohammadi et~al.}{2019}]{8965014}
\begin{bchapter}
\bauthor{\bsnm{Mohammadi}, \binits{S.}},
\bauthor{\bsnm{Majelan}, \binits{S.G.}},
\bauthor{\bsnm{Shokouhi}, \binits{S.B.}}:
\bctitle{Ensembles of deep neural networks for action recognition in still
  images}.
In: \bbtitle{2019 9th International Conference on Computer and Knowledge
  Engineering (ICCKE)},
pp. \bfpage{315}--\blpage{318}
(\byear{2019}).
\doiurl{10.1109/ICCKE48569.2019.8965014}
\end{bchapter}
\endbibitem

\bibitem[\protect\citeauthoryear{Ashrafi et~al.}{2021}]{AshrafiSeyed}
\begin{botherref}
\oauthor{\bsnm{Ashrafi}, \binits{S.S.}},
\oauthor{\bsnm{Shokouhi}, \binits{S.}},
\oauthor{\bsnm{Ayatollahi}, \binits{A.}}:
Action recognition in still images using a multi-attention guided network with
  weakly supervised saliency detection.
Multimedia Tools and Applications
\textbf{80}
(2021)
\doiurl{10.1007/s11042-021-11215-1}
\end{botherref}
\endbibitem

\bibitem[\protect\citeauthoryear{Zheng et~al.}{2020}]{Chen2004}
\begin{barticle}
\bauthor{\bsnm{Zheng}, \binits{Y.}},
\bauthor{\bsnm{Zheng}, \binits{X.}},
\bauthor{\bsnm{Lu}, \binits{X.}},
\bauthor{\bsnm{Wu}, \binits{S.}}:
\batitle{Spatial attention based visual semantic learning for action
  recognition in still images}.
\bjtitle{Neurocomputing}
\bvolume{413},
\bfpage{383}--\blpage{396}
(\byear{2020})
\doiurl{10.1016/j.neucom.2020.07.016}
\end{barticle}
\endbibitem

\bibitem[\protect\citeauthoryear{He et~al.}{2016}]{7780459}
\begin{bchapter}
\bauthor{\bsnm{He}, \binits{K.}},
\bauthor{\bsnm{Zhang}, \binits{X.}},
\bauthor{\bsnm{Ren}, \binits{S.}},
\bauthor{\bsnm{Sun}, \binits{J.}}:
\bctitle{Deep residual learning for image recognition}.
In: \bbtitle{2016 IEEE Conference on Computer Vision and Pattern Recognition
  (CVPR)},
pp. \bfpage{770}--\blpage{778}
(\byear{2016}).
\doiurl{10.1109/CVPR.2016.90}
\end{bchapter}
\endbibitem

\bibitem[\protect\citeauthoryear{Simonyan and Zisserman}{2014}]{51456465465465}
\begin{botherref}
\oauthor{\bsnm{Simonyan}, \binits{K.}},
\oauthor{\bsnm{Zisserman}, \binits{A.}}:
Very deep convolutional networks for large-scale image recognition.
arXiv 1409.1556
(2014)
\end{botherref}
\endbibitem

\bibitem[\protect\citeauthoryear{Oquab et~al.}{2014}]{6909618}
\begin{bchapter}
\bauthor{\bsnm{Oquab}, \binits{M.}},
\bauthor{\bsnm{Bottou}, \binits{L.}},
\bauthor{\bsnm{Laptev}, \binits{I.}},
\bauthor{\bsnm{Sivic}, \binits{J.}}:
\bctitle{Learning and transferring mid-level image representations using
  convolutional neural networks}.
In: \bbtitle{2014 IEEE Conference on Computer Vision and Pattern Recognition},
pp. \bfpage{1717}--\blpage{1724}
(\byear{2014}).
\doiurl{10.1109/CVPR.2014.222}
\end{bchapter}
\endbibitem

\bibitem[\protect\citeauthoryear{Vaswani et~al.}{2017}]{NIPS2017_3f5ee243}
\begin{bchapter}
\bauthor{\bsnm{Vaswani}, \binits{A.}},
\bauthor{\bsnm{Shazeer}, \binits{N.}},
\bauthor{\bsnm{Parmar}, \binits{N.}},
\bauthor{\bsnm{Uszkoreit}, \binits{J.}},
\bauthor{\bsnm{Jones}, \binits{L.}},
\bauthor{\bsnm{Gomez}, \binits{A.N.}},
\bauthor{\bsnm{Kaiser}, \binits{L.u.}},
\bauthor{\bsnm{Polosukhin}, \binits{I.}}:
\bctitle{Attention is all you need}.
In: \beditor{\bsnm{Guyon}, \binits{I.}},
\beditor{\bsnm{Luxburg}, \binits{U.V.}},
\beditor{\bsnm{Bengio}, \binits{S.}},
\beditor{\bsnm{Wallach}, \binits{H.}},
\beditor{\bsnm{Fergus}, \binits{R.}},
\beditor{\bsnm{Vishwanathan}, \binits{S.}},
\beditor{\bsnm{Garnett}, \binits{R.}} (eds.)
\bbtitle{Advances in Neural Information Processing Systems},
vol. \bseriesno{30}.
\bpublisher{Curran Associates, Inc.}, \blocation{???}
(\byear{2017}).
\burl{https://proceedings.neurips.cc/paper/2017/file/3f5ee243547dee91fbd053c1c4a845aa-Paper.pdf}
\end{bchapter}
\endbibitem

\bibitem[\protect\citeauthoryear{}{2021}]{vitjkljlfsd}
\begin{botherref}
An image is worth 16x16 words: Transformers for image recognition at scale.
arXiv:2010.11929v2
(2021)
\doiurl{10.48550/arXiv.2010.11929}
\end{botherref}
\endbibitem

\bibitem[\protect\citeauthoryear{Yan et~al.}{2018}]{8214269}
\begin{barticle}
\bauthor{\bsnm{Yan}, \binits{S.}},
\bauthor{\bsnm{Smith}, \binits{J.S.}},
\bauthor{\bsnm{Lu}, \binits{W.}},
\bauthor{\bsnm{Zhang}, \binits{B.}}:
\batitle{Multibranch attention networks for action recognition in still
  images}.
\bjtitle{IEEE Transactions on Cognitive and Developmental Systems}
\bvolume{10}(\bissue{4}),
\bfpage{1116}--\blpage{1125}
(\byear{2018})
\doiurl{10.1109/TCDS.2017.2783944}
\end{barticle}
\endbibitem

\bibitem[\protect\citeauthoryear{Thurau and Hlavac}{2008}]{Thurau2008PosePB}
\begin{botherref}
\oauthor{\bsnm{Thurau}, \binits{C.}},
\oauthor{\bsnm{Hlavac}, \binits{V.}}:
Pose primitive based human action recognition in videos or still images.
2008 IEEE Conference on Computer Vision and Pattern Recognition,
1--8
(2008)
\end{botherref}
\endbibitem

\bibitem[\protect\citeauthoryear{Bourdev and Malik}{2009}]{5459303}
\begin{bchapter}
\bauthor{\bsnm{Bourdev}, \binits{L.}},
\bauthor{\bsnm{Malik}, \binits{J.}}:
\bctitle{Poselets: Body part detectors trained using 3d human pose
  annotations}.
In: \bbtitle{2009 IEEE 12th International Conference on Computer Vision},
pp. \bfpage{1365}--\blpage{1372}
(\byear{2009}).
\doiurl{10.1109/ICCV.2009.5459303}
\end{bchapter}
\endbibitem

\bibitem[\protect\citeauthoryear{Gkioxari et~al.}{2015}]{7410641}
\begin{bchapter}
\bauthor{\bsnm{Gkioxari}, \binits{G.}},
\bauthor{\bsnm{Girshick}, \binits{R.}},
\bauthor{\bsnm{Malik}, \binits{J.}}:
\bctitle{Actions and attributes from wholes and parts}.
In: \bbtitle{2015 IEEE International Conference on Computer Vision (ICCV)},
pp. \bfpage{2470}--\blpage{2478}
(\byear{2015}).
\doiurl{10.1109/ICCV.2015.284}
\end{bchapter}
\endbibitem

\bibitem[\protect\citeauthoryear{Newell et~al.}{2016}]{Hourglass}
\begin{botherref}
\oauthor{\bsnm{Newell}, \binits{A.}},
\oauthor{\bsnm{Yang}, \binits{K.}},
\oauthor{\bsnm{Deng}, \binits{J.}}:
Stacked hourglass networks for human pose estimation.
ECCV 2016. ECCV 2016. Lecture Notes in Computer Science
\textbf{9912}
(2016)
\end{botherref}
\endbibitem

\bibitem[\protect\citeauthoryear{Li et~al.}{2020}]{LI2020107341}
\begin{barticle}
\bauthor{\bsnm{Li}, \binits{Y.}},
\bauthor{\bsnm{Li}, \binits{K.}},
\bauthor{\bsnm{Wang}, \binits{X.}}:
\batitle{Recognizing actions in images by fusing multiple body structure cues}.
\bjtitle{Pattern Recognition}
\bvolume{104},
\bfpage{107341}
(\byear{2020})
\doiurl{10.1016/j.patcog.2020.107341}
\end{barticle}
\endbibitem

\bibitem[\protect\citeauthoryear{Yao and Fei-Fei}{2010}]{5540235}
\begin{bchapter}
\bauthor{\bsnm{Yao}, \binits{B.}},
\bauthor{\bsnm{Fei-Fei}, \binits{L.}}:
\bctitle{Modeling mutual context of object and human pose in human-object
  interaction activities}.
In: \bbtitle{2010 IEEE Computer Society Conference on Computer Vision and
  Pattern Recognition},
pp. \bfpage{17}--\blpage{24}
(\byear{2010}).
\doiurl{10.1109/CVPR.2010.5540235}
\end{bchapter}
\endbibitem

\bibitem[\protect\citeauthoryear{Desai et~al.}{2010}]{5543176}
\begin{bchapter}
\bauthor{\bsnm{Desai}, \binits{C.}},
\bauthor{\bsnm{Ramanan}, \binits{D.}},
\bauthor{\bsnm{Fowlkes}, \binits{C.}}:
\bctitle{Discriminative models for static human-object interactions}.
In: \bbtitle{2010 IEEE Computer Society Conference on Computer Vision and
  Pattern Recognition - Workshops},
pp. \bfpage{9}--\blpage{16}
(\byear{2010}).
\doiurl{10.1109/CVPRW.2010.5543176}
\end{bchapter}
\endbibitem

\bibitem[\protect\citeauthoryear{Gkioxari
  et~al.}{2017}]{Gkioxari2017DetectingAR}
\begin{botherref}
\oauthor{\bsnm{Gkioxari}, \binits{G.}},
\oauthor{\bsnm{Girshick}, \binits{R.B.}},
\oauthor{\bsnm{Doll{\'a}r}, \binits{P.}},
\oauthor{\bsnm{He}, \binits{K.}}:
Detecting and recognizing human-object interactions.
2018 IEEE/CVF Conference on Computer Vision and Pattern Recognition,
8359--8367
(2017)
\end{botherref}
\endbibitem

\bibitem[\protect\citeauthoryear{Gkioxari et~al.}{2015}]{7410486}
\begin{bchapter}
\bauthor{\bsnm{Gkioxari}, \binits{G.}},
\bauthor{\bsnm{Girshick}, \binits{R.}},
\bauthor{\bsnm{Malik}, \binits{J.}}:
\bctitle{Contextual action recognition with r* cnn}.
In: \bbtitle{2015 IEEE International Conference on Computer Vision (ICCV)},
pp. \bfpage{1080}--\blpage{1088}
(\byear{2015}).
\doiurl{10.1109/ICCV.2015.129}
\end{bchapter}
\endbibitem

\bibitem[\protect\citeauthoryear{Delaitre et~al.}{2010}]{BMVC.24.97}
\begin{bchapter}
\bauthor{\bsnm{Delaitre}, \binits{V.}},
\bauthor{\bsnm{Laptev}, \binits{I.}},
\bauthor{\bsnm{Sivic}, \binits{J.}}:
\bctitle{Recognizing human actions in still images: a study of bag-of-features
  and part-based representations}.
In: \bbtitle{Proceedings of the British Machine Vision Conference},
pp. \bfpage{97}--\blpage{19711}.
\bpublisher{BMVA Press}, \blocation{???}
(\byear{2010}).
\bcomment{doi:10.5244/C.24.97}
\end{bchapter}
\endbibitem

\bibitem[\protect\citeauthoryear{Zhang et~al.}{2016}]{7558119}
\begin{barticle}
\bauthor{\bsnm{Zhang}, \binits{Y.}},
\bauthor{\bsnm{Cheng}, \binits{L.}},
\bauthor{\bsnm{Wu}, \binits{J.}},
\bauthor{\bsnm{Cai}, \binits{J.}},
\bauthor{\bsnm{Do}, \binits{M.N.}},
\bauthor{\bsnm{Lu}, \binits{J.}}:
\batitle{Action recognition in still images with minimum annotation efforts}.
\bjtitle{IEEE Transactions on Image Processing}
\bvolume{25}(\bissue{11}),
\bfpage{5479}--\blpage{5490}
(\byear{2016})
\doiurl{10.1109/TIP.2016.2605305}
\end{barticle}
\endbibitem

\bibitem[\protect\citeauthoryear{Mnih et~al.}{2014}]{article524454}
\begin{botherref}
\oauthor{\bsnm{Mnih}, \binits{V.}},
\oauthor{\bsnm{Heess}, \binits{N.}},
\oauthor{\bsnm{Graves}, \binits{A.}},
\oauthor{\bsnm{Kavukcuoglu}, \binits{K.}}:
Recurrent models of visual attention.
Advances in Neural Information Processing Systems
\textbf{3}
(2014)
\end{botherref}
\endbibitem

\bibitem[\protect\citeauthoryear{Bahdanau et~al.}{2014}]{DzmitryBahdanau}
\begin{botherref}
\oauthor{\bsnm{Bahdanau}, \binits{D.}},
\oauthor{\bsnm{Cho}, \binits{K.}},
\oauthor{\bsnm{Bengio}, \binits{Y.}}:
Neural machine translation by jointly learning to align and translate.
Advances in Neural Information Processing Systems
(2014)
\end{botherref}
\endbibitem

\bibitem[\protect\citeauthoryear{He et~al.}{2021}]{9586824}
\begin{bchapter}
\bauthor{\bsnm{He}, \binits{W.}},
\bauthor{\bsnm{Wu}, \binits{Y.}},
\bauthor{\bsnm{Li}, \binits{X.}}:
\bctitle{Attention mechanism for neural machine translation: A survey}.
In: \bbtitle{2021 IEEE 5th Information Technology,Networking,Electronic and
  Automation Control Conference (ITNEC)},
vol. \bseriesno{5},
pp. \bfpage{1485}--\blpage{1489}
(\byear{2021}).
\doiurl{10.1109/ITNEC52019.2021.9586824}
\end{bchapter}
\endbibitem

\bibitem[\protect\citeauthoryear{Xu et~al.}{2015}]{Gkioxari2015ContextualLM}
\begin{botherref}
\oauthor{\bsnm{Xu}, \binits{K.}},
\oauthor{\bsnm{Ba}, \binits{J.}},
\oauthor{\bsnm{Kiros}, \binits{R.}},
\oauthor{\bsnm{Cho}, \binits{K.}},
\oauthor{\bsnm{Courville}, \binits{A.}},
\oauthor{\bsnm{Salakhutdinov}, \binits{R.}},
\oauthor{\bsnm{Zemel}, \binits{R.}},
\oauthor{\bsnm{Bengio}}:
Show, attend and tell: Neural image caption generation with visual attention
(2015)
\end{botherref}
\endbibitem

\bibitem[\protect\citeauthoryear{Zhao et~al.}{2017}]{7807286}
\begin{barticle}
\bauthor{\bsnm{Zhao}, \binits{B.}},
\bauthor{\bsnm{Wu}, \binits{X.}},
\bauthor{\bsnm{Feng}, \binits{J.}},
\bauthor{\bsnm{Peng}, \binits{Q.}},
\bauthor{\bsnm{Yan}, \binits{S.}}:
\batitle{Diversified visual attention networks for fine-grained object
  classification}.
\bjtitle{IEEE Transactions on Multimedia}
\bvolume{19}(\bissue{6}),
\bfpage{1245}--\blpage{1256}
(\byear{2017})
\doiurl{10.1109/TMM.2017.2648498}
\end{barticle}
\endbibitem

\bibitem[\protect\citeauthoryear{Ashrafi et~al.}{2023}]{ashrafi2023}
\begin{barticle}
\bauthor{\bsnm{Ashrafi}, \binits{S.S.}},
\bauthor{\bsnm{Shokouhi}, \binits{S.B.}},
\bauthor{\bsnm{Ayatollahi}, \binits{A.}}:
\batitle{Still image action recognition based on interactions between joints
  and objects}.
\bjtitle{Multimedia Tools and Applications}
\bvolume{82},
\bfpage{25945}--\blpage{25971}
(\byear{2023})
\doiurl{10.1007/s11042-023-14350-z}
\end{barticle}
\endbibitem

\bibitem[\protect\citeauthoryear{Bas and Ikizler-Cinbis}{2022}]{BAS2022116664}
\begin{barticle}
\bauthor{\bsnm{Bas}, \binits{C.}},
\bauthor{\bsnm{Ikizler-Cinbis}, \binits{N.}}:
\batitle{Top-down and bottom-up attentional multiple instance learning for
  still image action recognition}.
\bjtitle{Signal Processing: Image Communication}
\bvolume{104},
\bfpage{116664}
(\byear{2022})
\doiurl{10.1016/j.image.2022.116664}
\end{barticle}
\endbibitem

\bibitem[\protect\citeauthoryear{Ren et~al.}{2015}]{NIPS2015_14bfa6bb}
\begin{bchapter}
\bauthor{\bsnm{Ren}, \binits{S.}},
\bauthor{\bsnm{He}, \binits{K.}},
\bauthor{\bsnm{Girshick}, \binits{R.}},
\bauthor{\bsnm{Sun}, \binits{J.}}:
\bctitle{Faster r-cnn: Towards real-time object detection with region proposal
  networks}.
In: \beditor{\bsnm{Cortes}, \binits{C.}},
\beditor{\bsnm{Lawrence}, \binits{N.}},
\beditor{\bsnm{Lee}, \binits{D.}},
\beditor{\bsnm{Sugiyama}, \binits{M.}},
\beditor{\bsnm{Garnett}, \binits{R.}} (eds.)
\bbtitle{Advances in Neural Information Processing Systems},
vol. \bseriesno{28}.
\bpublisher{Curran Associates, Inc.}, \blocation{???}
(\byear{2015}).
\burl{https://proceedings.neurips.cc/paper/2015/file/14bfa6bb14875e45bba028a21ed38046-Paper.pdf}
\end{bchapter}
\endbibitem

\bibitem[\protect\citeauthoryear{Liu et~al.}{2018}]{8579011}
\begin{bchapter}
\bauthor{\bsnm{Liu}, \binits{S.}},
\bauthor{\bsnm{Qi}, \binits{L.}},
\bauthor{\bsnm{Qin}, \binits{H.}},
\bauthor{\bsnm{Shi}, \binits{J.}},
\bauthor{\bsnm{Jia}, \binits{J.}}:
\bctitle{Path aggregation network for instance segmentation}.
In: \bbtitle{2018 IEEE/CVF Conference on Computer Vision and Pattern
  Recognition},
pp. \bfpage{8759}--\blpage{8768}
(\byear{2018}).
\doiurl{10.1109/CVPR.2018.00913}
\end{bchapter}
\endbibitem

\bibitem[\protect\citeauthoryear{Yao et~al.}{2011}]{VALLIM20136258fdsfdsfsdf}
\begin{botherref}
\oauthor{\bsnm{Yao}, \binits{B.}},
\oauthor{\bsnm{Jiang}, \binits{X.}},
\oauthor{\bsnm{Khosla}, \binits{A.}},
\oauthor{\bsnm{Lin}, \binits{A.L.}},
\oauthor{\bsnm{Guibas}, \binits{L.J.}},
\oauthor{\bsnm{Fei-Fei}, \binits{L.}}:
Human action recognition by learning bases of action attributes and parts.
Internation Conference on Computer Vision (ICCV)
(2011)
\end{botherref}
\endbibitem

\bibitem[\protect\citeauthoryear{Everingham et~al.}{2010}]{EveringhamGWWZ10}
\begin{barticle}
\bauthor{\bsnm{Everingham}, \binits{M.}},
\bauthor{\bsnm{Gool}, \binits{L.V.}},
\bauthor{\bsnm{Williams}, \binits{C.K.I.}},
\bauthor{\bsnm{Winn}, \binits{J.M.}},
\bauthor{\bsnm{Zisserman}, \binits{A.}}:
\batitle{The pascal visual object classes (voc) challenge.}
\bjtitle{Int. J. Comput. Vis.}
\bvolume{88}(\bissue{2}),
\bfpage{303}--\blpage{338}
(\byear{2010})
\end{barticle}
\endbibitem

\bibitem[\protect\citeauthoryear{Zhang et~al.}{2018}]{mixup}
\begin{botherref}
\oauthor{\bsnm{Zhang}, \binits{H.}},
\oauthor{\bsnm{Cisse}, \binits{M.}},
\oauthor{\bsnm{Dauphin}, \binits{Y.N.}},
\oauthor{\bsnm{Lopez-Paz}, \binits{D.}}:
mixup: Beyond empirical risk minimization
(2018)
\doiurl{10.48550/arXiv.1710.09412}
\end{botherref}
\endbibitem

\bibitem[\protect\citeauthoryear{Cubuk et~al.}{2019}]{RandAugment}
\begin{botherref}
\oauthor{\bsnm{Cubuk}, \binits{E.D.}},
\oauthor{\bsnm{Zoph}, \binits{B.}},
\oauthor{\bsnm{Shlens}, \binits{J.}},
\oauthor{\bsnm{Le}, \binits{Q.V.}}:
Randaugment: Practical automated data augmentation with a reduced search space
(2019)
\doiurl{10.48550/arXiv.1909.13719}
\end{botherref}
\endbibitem

\bibitem[\protect\citeauthoryear{Russakovsky et~al.}{2015}]{ImageNet}
\begin{botherref}
\oauthor{\bsnm{Russakovsky}, \binits{O.}},
\oauthor{\bsnm{Deng}, \binits{J.}},
\oauthor{\bsnm{Su}, \binits{H.}},
\oauthor{\bsnm{Krause}, \binits{J.}},
\oauthor{\bsnm{Satheesh}, \binits{S.}},
\oauthor{\bsnm{Ma}, \binits{S.}},
\oauthor{\bsnm{Huang}, \binits{Z.}},
\oauthor{\bsnm{Karpathy}, \binits{A.}},
\oauthor{\bsnm{Khosla}, \binits{A.}},
\oauthor{\bsnm{Bernstein}, \binits{M.}},
\oauthor{\bsnm{Berg}, \binits{A.C.}},
\oauthor{\bsnm{Fei-Fei}, \binits{L.}}:
Imagenet large scale visual recognition challenge
(2015)
\doiurl{10.48550/arXiv.1409.0575}
\end{botherref}
\endbibitem

\bibitem[\protect\citeauthoryear{Paszke et~al.}{2019}]{NEURIPS2019_9015}
\begin{bchapter}
\bauthor{\bsnm{Paszke}, \binits{A.}},
\bauthor{\bsnm{Gross}, \binits{S.}},
\bauthor{\bsnm{Massa}, \binits{F.}},
\bauthor{\bsnm{Lerer}, \binits{A.}},
\bauthor{\bsnm{Bradbury}, \binits{J.}},
\bauthor{\bsnm{Chanan}, \binits{G.}},
\bauthor{\bsnm{Killeen}, \binits{T.}},
\bauthor{\bsnm{Lin}, \binits{Z.}},
\bauthor{\bsnm{Gimelshein}, \binits{N.}},
\bauthor{\bsnm{Antiga}, \binits{L.}},
\bauthor{\bsnm{Desmaison}, \binits{A.}},
\bauthor{\bsnm{Kopf}, \binits{A.}},
\bauthor{\bsnm{Yang}, \binits{E.}},
\bauthor{\bsnm{DeVito}, \binits{Z.}},
\bauthor{\bsnm{Raison}, \binits{M.}},
\bauthor{\bsnm{Tejani}, \binits{A.}},
\bauthor{\bsnm{Chilamkurthy}, \binits{S.}},
\bauthor{\bsnm{Steiner}, \binits{B.}},
\bauthor{\bsnm{Fang}, \binits{L.}},
\bauthor{\bsnm{Bai}, \binits{J.}},
\bauthor{\bsnm{Chintala}, \binits{S.}}:
\bctitle{Pytorch: An imperative style, high-performance deep learning library}.
In: \bbtitle{Advances in Neural Information Processing Systems 32},
pp. \bfpage{8024}--\blpage{8035}.
\bpublisher{Curran Associates, Inc.}, \blocation{???}
(\byear{2019}).
\burl{http://papers.neurips.cc/paper/9015-pytorch-an-imperative-style-high-performance-deep-learning-library.pdf}
\end{bchapter}
\endbibitem

\bibitem[\protect\citeauthoryear{Ma and Liang}{2020}]{9102933}
\begin{bchapter}
\bauthor{\bsnm{Ma}, \binits{W.}},
\bauthor{\bsnm{Liang}, \binits{S.}}:
\bctitle{Human-object relation network for action recognition in still images}.
In: \bbtitle{2020 IEEE International Conference on Multimedia and Expo (ICME)},
pp. \bfpage{1}--\blpage{6}
(\byear{2020}).
\doiurl{10.1109/ICME46284.2020.9102933}
\end{bchapter}
\endbibitem

\bibitem[\protect\citeauthoryear{Selvaraju et~al.}{2017}]{8237336}
\begin{bchapter}
\bauthor{\bsnm{Selvaraju}, \binits{R.R.}},
\bauthor{\bsnm{Cogswell}, \binits{M.}},
\bauthor{\bsnm{Das}, \binits{A.}},
\bauthor{\bsnm{Vedantam}, \binits{R.}},
\bauthor{\bsnm{Parikh}, \binits{D.}},
\bauthor{\bsnm{Batra}, \binits{D.}}:
\bctitle{Grad-cam: Visual explanations from deep networks via gradient-based
  localization}.
In: \bbtitle{2017 IEEE International Conference on Computer Vision (ICCV)},
pp. \bfpage{618}--\blpage{626}
(\byear{2017}).
\doiurl{10.1109/ICCV.2017.74}
\end{bchapter}
\endbibitem

\end{thebibliography}

\end{document}